\documentclass[10pt]{article} 
\usepackage[preprint]{tmlr}

\usepackage{amsmath,amsfonts,bm}

\def\eqref#1{equation~\ref{#1}}

\def\1{\bm{1}}

\DeclareMathAlphabet{\mathsfit}{\encodingdefault}{\sfdefault}{m}{sl}
\SetMathAlphabet{\mathsfit}{bold}{\encodingdefault}{\sfdefault}{bx}{n}

\DeclareMathOperator*{\argmin}{arg\,min}

\usepackage{hyperref}
\usepackage{url}
\usepackage{graphicx}
\usepackage{natbib}
\usepackage{algorithm}
\usepackage{algpseudocode}
\usepackage{booktabs}
\usepackage{amsmath}
\usepackage[utf8]{inputenc} 
\usepackage[T1]{fontenc}    
\usepackage{amsfonts}       
\usepackage{nicefrac}       
\usepackage{microtype}      
\usepackage{xcolor}         
\usepackage{amssymb}
\usepackage{mathtools}
\usepackage{amsthm}
\usepackage{arydshln}
\usepackage{multirow}

\title{Optimal Hydra: Model Heterogeneous Federated Learning for LLMs using Data-Informed Hydra-Nets}

\author{%
\centering
  \begin{tabular}[t]{c@{\hskip 2.5em}c@{\hskip 2.5em}c}
    \name Gwen Legate & \name Rahaf Aljundi & \name Eugene Belilovsky \\
    \addr Concordia University & \addr Toyota Motor Europe & \addr Concordia University  \\
    \addr Mila & & \addr Mila
  \end{tabular}
}

\def\month{MM}  
\def\year{YYYY} 
\def\openreview{\url{https://openreview.net/forum?id=XXXX}} 

\begin{document}

\maketitle

\begin{abstract}
Large language models (LLMs) trained on vast amounts of data have demonstrated impressive performance over a wide range of tasks, but future performance gains will require increasing amounts of data and compute. Federated Learning (FL) can facilitate training on decentralized private data, but client resource limitations can make FL training impractical, particularly as model sizes continue to expand. These resource constraints make the widely accepted assumption of a shared server–client architecture impractical. Model heterogeneous FL offers a promising alternative by adapting model size to client capabilities; however, previous work is limited to the pre-training setting, and these conclusions on sub-network selection do not necessarily hold for fine-tuning. Furthermore, prior approaches have created client sub-networks without considering their unique data distributions, thereby overlooking the opportunity to tailor sub-network architecture to local data. Motivated by evidence that attention heads exhibit task dependency within layer defined feature spaces, we propose Optimal Hydra, a transformer based heterogeneous FL framework for fine-tuning that combines global layer selection with client specific attention head selection aligned with local data distributions. Optimal Hydra makes explicit use of client data to produce client-specific sub-networks we call Hydra-nets. To our knowledge, this is the first model heterogeneous FL approach tailored to LLM fine-tuning. Experiments on three datasets show significant improvements over a number of baseline methods while also reducing local resource demands.
\end{abstract}

\section{Introduction}
Foundation models have demonstrated advanced abilities on diverse tasks and have been widely adopted in domains including healthcare, education and autonomous vehicles \citep{bommasani2021opportunities, thirunavukarasu2023large, brown2020language, hu2023planning}. Established scaling laws indicate that continued performance gains will require increasingly large amounts of data \citep{kaplan2020scaling}, a requirement at odds with the growing awareness of digital privacy. Federated learning (FL) \citep{mcmahan2017communication} offers a privacy preserving alternative that may create a greater willingness to share data \citet{sani2024future}; however, the application of FL to foundation models is challenging due to large parameter counts that considerably exceed the capabilities of a typical edge device \citep{yao2024survey}. Previous attempts to bridge this gap have largely relied on Parameter-Efficient Fine-Tuning (PEFT) \citep{babakniya2023slora, tian2024hydralora, bai2024federated}, but these methods can struggle to match the performance of full fine-tuning and are prone to overfitting when parameters are strictly limited \citep{chen2022revisiting}. Furthermore, these methods still require homogeneous model architectures, so client nodes must have adequate resources to store and train the entire large model, which is often not the case \citep{yao2024survey}. This creates a dilemma: either we limit model complexity to accommodate the most under resourced client or we exclude low resource clients from training. Neither of these options is ideal, since limiting model complexity eliminates our ability to train large SoTA models and excluding low resource clients from training creates a bias towards the data of those with better resources \citep{kairouz2021advances}.

\noindent Model heterogeneous FL relaxes the assumption that server and client model architectures must be identical, it allows for federated training with heterogeneous models better suited to unique client capabilities. Early work used static \citep{diaoheterofl, horvath2021fjord} or random \citep{wen2022feddrop} sub-networks that trained model parameters unequally, making them susceptible to client drift \citep{karimireddy2020scaffold}. Subsequent work improved stability by training model parameters more equally \citep{alam2022fedrolex, tianchi2025fedbrb}; however, all of these approaches overlook the fact that optimal sub-networks are intrinsically linked to the data they encounter \citep{franklelottery}. For FL in particular, client data distributions are often highly heterogeneous and  tailoring sub-network architectures to local data presents an important opportunity to capture distribution-specific features. Notably, while prior work focuses exclusively on the pre-training setting, our work addresses fine-tuning and we will show that previous conclusions regarding strictly equal parameter training in the pre-training setting do not necessarily hold in the context of fine-tuning.

\begin{figure*}
    \centering
    \includegraphics[width=0.6\textwidth]{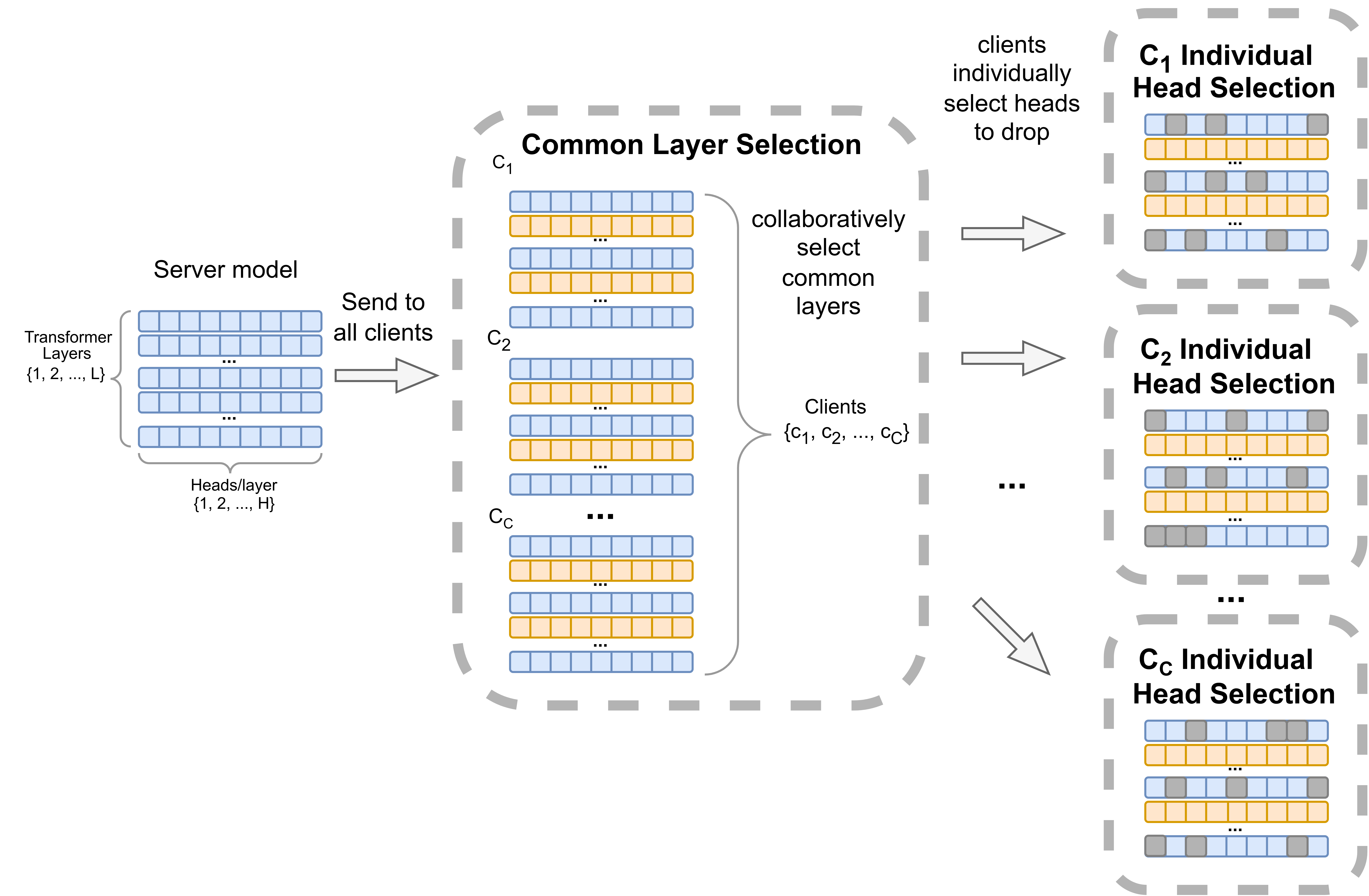}
    \caption{Each model is represented as a set of horizontal rectangles denoting transformer layers. The segregation of each layer represents attention heads present in the layer. Note: for simplicity, attention heads are the only architectural components depicted. \textbf{Hydra-Net selection algorithm:} Clients first collaboratively select a subset of common layers from which no attention heads may be removed; these layers are highlighted in orange to indicate the protected status of their attention heads during the individual head selection process. Next, clients individually select a subset of heads in each unimportant layer via a gradient based scoring mechanism. Heads that were not selected by clients in the individual selection process are grey to indicate their removal from the sub-network. The end result is a unique, data aligned Hydra-Net for each client that is subsequently used in the model heterogeneous FL framework, Optimal Hydra.}
    \label{fig:subnet_select}
\end{figure*}

\noindent We propose \textit{Optimal Hydra}, a method that adapts model-heterogeneous FL for LLM fine-tuning. Prior research has focused exclusively on pre-training \citep{alam2022fedrolex, diaoheterofl, tianchi2025fedbrb, wen2022feddrop, horvath2021fjord}; we address this gap in the literature by being the first to investigate model-heterogeneous FL in the fine-tuning setting, a setting that has become increasingly relevant as large foundation models become the norm. We also take the opportunity to improve upon the random sub-network selection used in previous works by creating data-informed sub-networks we call \textit{Hydra-Nets} that are tailored to local client data to better capture distribution-specific features. Fig.~\ref{fig:subnet_select} provides a high level overview of our Hydra-Net selection method that proceeds in two steps; \textit{common layer selection} and \textit{individual head selection}. Common layer selection is motivated by the fact that transformer layers govern macro level feature transformations \citep{voita2019analyzing} and that in FL a lack of consensus amongst client sub-networks can negatively affect global model convergence \citep{diaoheterofl, horvath2021fjord, babakniya2023revisiting}. We adapt a metric used to determine the relative importance of transformer layers  by \citet{he2024notallattention} to enable clients to collaboratively identify a small subset of ``important" layers, establishing a shared structural backbone across clients from which no attention heads may be removed. Subsequently, each individual client will select a unique subset of attention heads to keep from each ``unimportant" layer via a gradient based scoring metric, the rest will be discarded. Ultimately, Hydra-Net selection is motivated by the task dependent nature of attention heads to create targeted sub-networks, informed by each client's unique local data distributions.\\

\noindent Our key contributions are as follows:
 \begin{itemize}
    \item We propose Optimal Hydra for model heterogeneous federated fine-tuning using data-informed LLM sub-networks.As the first model-heterogeneous FL method to tailor sub-network architectures to local client data, Optimal Hydra achieves up to a $42.5\%$ improvement over the best performing baseline in challenging model heterogeneous conditions and achieves reductions in computational and memory resource utilization up to $32.0\%$ and $29.3\%$, respectively.
    \item We demonstrate that assumptions from the pre-training about evenly distributing parameters can be relaxed in the fine-tuning regime due to the strong initial representations provided by the pre-trained model.
    \item Based on representation dynamics and prior work suggesting that attention heads may be linked to specific tasks within layer feature spaces, we combine common layer selection to create Hydra-Nets, data-informed sub-networks that are better able to capture distribution specific features.
 \end{itemize}

\section{Background and Related Work}
Federated learning (FL) \citep{mcmahan2017communication} is a privacy preserving, distributed, machine learning paradigm that allows collaborative model training without data sharing.  Resource limitations on federated edge devices can make FL challenging to execute using current SotA foundation models, examples in the literature \citep{babakniya2023slora, bai2024federated, tian2024hydralora, wu2024fedbiot}  have relied heavily on parameter efficient fine-tuning (PEFT) which significantly reduces communication costs but remains highly burdensome in terms of local computation and memory requirements. While communication constraints have historically been considered the most significant federated resource bottleneck \citep{mcmahan2017communication, kairouz2021advances}, the large parameter counts of modern foundation models have increased the significance of memory and computational resource bottlenecks \citep{yao2024survey, sani2024future} since the common assumption in FL, including the aforementioned PEFT methods, is that server and client models possess the same architecture. This underlying assumption either limits the choice of global model so that the most resource limited client is able to participate; or selects a more complex global model and excludes resource limited clients altogether. A natural solution is to distribute sub-networks according to client capabilities, but stable aggregation of heterogeneous local architectures into a single global model is challenging and requires careful consideration of base and sub-network model architectures in order to be successful \citep{alam2022fedrolex}.  

\noindent HeteroFL \citep{diaoheterofl}, widely regarded as the first example of model heterogeneous FL in the literature, creates static sub-models that train model parameters unequally. FedDrop \citep{wen2022feddrop} randomly drops a fraction of neurons in each layer, which can lead to inconsistent updates and convergence instability, particularly when data are not independently and identically distributed (i.i.d.). FedRolex \citep{alam2022fedrolex} and FedBRB \citep{tianchi2025fedbrb} improve upon these methods by iterating through model parameters so they are all trained evenly, yet these approaches fail to consider the quality of the sub-models and how differences in local data distributions can affect the suitability of the selected sub-network from round to round. These approaches are designed for pre-training, use relatively small models and focus primarily on computer vision tasks. Due to the learning differences that exist between federated pre-training and fine-tuning \citep{nguyenbegin, legate2023guiding}, we cannot assume their suitability for large model fine-tuning. Additionally, pervious model heterogeneous FL experiments have used convolutional neural networks (CNNs), none of them have been adapted for the transformer architecture, ubiquitous in modern machine learning. 

\noindent The opacity of an LLMs internal mechanisms makes improving the efficiency of supervised fine-tuning (SFT) methods challenging. Previous attention head focused interpretability studies suggest that they exhibit a high degree of task specificity \citep{zhao2024supervised, yin2024lofit} and redundancy \citep{michel2019sixteen, diaoheterofl} during the SFT process. \citet{zhao2024supervised} observe rapid task adaptation of attention heads during SFT, they also show task dependence by targeting specific heads, which consistently provides better performance. LoFiT \citep{yin2024lofit} trains small localized adapters to fine tune attention heads. The authors find that identifying and targeting task heads most important for a task is more effective than standard fine-tuning, random head selection or targeting specific layers identified by probing. \citet{michel2019sixteen} find that while many heads are task dependent, some are universally important regardless of task. 

\section{Optimal Hydra}
Optimal Hydra adapts model heterogeneous FL for transformer models using heterogeneous client sub-networks called Hydra-Nets, clients train their Hydra-Nets locally and send updates back to the server for aggregation. While aggregating updates from heterogeneous local sub-networks into a single global model has presented a significant challenge for model heterogeneous FL, our approach overcomes this difficulty through direct index-mapping. During the aggregation phase, local updates are mapped back to their original coordinates in the global transformer architecture and parameter averaging is scaled by the total number of data samples residing in clients contributing to the update of a given parameter instead of the typical FL approach of scaling updates according to the number of data samples residing in all clients participating in the federated round \citep{mcmahan2017communication}. This approach lets the global model accumulate specialized, data-informed gradients from specific heads without the gradient interference present from full parameter federated averaging.

\subsection{Problem Formation}
\label{sec:formulation}
In FL, distributed optimization occurs over $C$ clients where each client $c\in\{1, ..., C\}$ possesses data $\mathbf{X}_{c},\mathbf{Y}_{c}$ containing $n_c$ samples drawn from distribution $D_c$. We define the total number of samples across all clients as $n=\sum_{c=1}^{C}n_{c}$. The data $\mathbf{X}_{c}$ at each node may be drawn from different distributions and/or may be unbalanced with some clients possessing more training samples than others. The typical objective function for federated optimization is given by Eq.~\ref{eq:fl_objective} \citep{mcmahan2017communication} where $\theta\in\mathbb{R}^d$ are the model weights and $\mathcal{L}$ is the objective function. 

\begin{equation} \label{eq:fl_objective}
\mathbf{\theta}^*\in \argmin_{\mathbf{\theta}}\sum_{c=1}^{C}\frac{n_{c}}{n}\mathcal{L}(f(\mathbf{\theta}, \mathbf{X}_{c}))
\end{equation}

\noindent We undertake federated fine-tuning of a server-side LLM by a set of resource heterogeneous client nodes. Federated clients typically possess relatively small datasets and might not require an LLMs full capability for performance on unrelated domains. We define $\mathbf{\theta}_{c}\in\mathbb{R}^{sd}$ as an sd-dimensional set of Hydra-Net parameters for client $c\in C$ where $\forall c\in C\;sd_c\subsetneq d$ and modify Eq.~\ref{eq:fl_objective} as follows.

\begin{equation} \label{eq:our_fl_objective}
\mathbf{\theta}^*\in \argmin_{\mathbf{\theta}}\sum_{c=1}^{C}\frac{n_{c}}{n}\mathcal{L}(f(\mathbf{\theta_c}, \mathbf{X}_{c}))
\end{equation}

\noindent The relationship between Hydra-Nets for two clients $c1$ and $c2$ where $c1\neq c2$ is that they share at least the set of globally common layers, and can share all, none or some of the individually selected heads in the remaining layers.

\noindent Hydra-Net selection applies to multi-head attention, decoder only, transformer architecture with a set of heads, $H$ and layers, $L$ . In what follows, we will use $H_\ell^{(c)}$ to indicate the set of heads in layer $\ell\in L$, belonging to client $c\in C$, the absence of a superscript denotes the server model. Multi-head a
ttention involves the concatenation of multiple self-attention blocks computed using query, key and value matrices $\mathbf{W}_{qh}$, $\mathbf{W}_{kh}$ and $\mathbf{W}_{vh}$ for each head $h\in H$ \citep{vaswani2017attention}. 

\subsection{Hydra-Net Selection}
Model heterogeneity creates particular challenges for model aggregation in FL and these systems are particularly sensitive to client drift \citep{karimireddy2020scaffold} because of their un-equal parameter training \citep{alam2022fedrolex}. One of our key contributions is the use of data-informed sub-networks to stabilize aggregation and improve training outcomes. To this end, we combine a number of previous insights to optimize our Hydra-Net selection. Attention heads have been shown to exhibit a high degree of task specificity \citep{zhao2024supervised, yin2024lofit} and certain attention heads exhibit universal importance across domains \citet{michel2019sixteen}. Because these critical heads concentrate within key functional layers that drive core feature transformations \citep{voita2019analyzing} we establish a two stage Hydra-Net selection strategy: common layer selection and individual head selection. As shown in Fig.~\ref{fig:subnet_select}, clients first work collaboratively to select a small number of layers from which no attention heads are dropped. They subsequently select attention heads with high sensitivity to their own local data distributions, which they keep from the remaining layers while dropping the rest. In what follows, we present a detailed discussion of  the procedures for common and individual head selection.
\vspace{-10pt}

\subsubsection{Common Layer Selection}
\label{sec:common}
\begin{algorithm}
\caption{SelectCommonLayers}
\label{alg:common}
\begin{algorithmic}[1]
\Require Client set $\mathcal{C}$, local data $\{X_c\}_{c \in \mathcal{C}}$, pre-trained params $\theta = \{\theta_\ell\}_{\ell=1}^L$
\ForAll{$c \in \mathcal{C}$}
    \State $H^{(c)}_0 \gets X_c$
\EndFor
\For{$\ell = 1$ to $L$}
    \ForAll{$c \in \mathcal{C}$}
        \State $H^{(c)}_\ell \gets f(\theta_\ell, H^{(c)}_{\ell-1})$
        \State $cs_{c,\ell} \gets \text{cosine\_similarity}(H^{(c)}_{\ell-1}, H^{(c)}_\ell)$
    \EndFor
    \State $cs_\ell \gets \frac{1}{|\mathcal{C}|} \sum_{c \in \mathcal{C}} cs_{c,\ell}$
\EndFor
\State $SC \gets \{cs_\ell\}_{\ell=1}^L$
\State $\mathbf{z} \gets \text{GMM}(SC)$
\State $\mathcal{L}_{\text{keep}} \gets \{\ell \mid z_\ell \in \{0,1\}\}$
\State $\mathcal{L}_{\text{drop}} \gets \{\ell \mid z_\ell = 2\}$
\Return $\mathcal{L}_{\text{keep}}, \mathcal{L}_{\text{drop}}$
\end{algorithmic}
\end{algorithm}
\vspace{-4pt}

When developing sub-network selection strategies, it is tempting to consider individual client properties in isolation; however, this approach over simplifies the problem and a lack of consensus between client sub-networks can inhibit global model convergence and cause serious performance degradation \citep{babakniya2023revisiting, alam2022fedrolex}. To prevent this from occurring in Hydra-Nets, all clients retain a common subset of parameters. Instead of attempting to negotiate consensus on individual attention heads which risks over-specializing  or missing critical head interactions, we select common parameters by layer instead of opting for fine grained head-selection. \citet{he2024notallattention} use a metric based on cosine similarity to score the relative importance of transformer layers, where layers with high similarity between inputs and outputs are deemed redundant. This provides a computationally efficient proxy to identify layers that drive critical representation shifts. We evaluate layer redundancy locally at each client using activation similarity statistics of each transformer block (Alg.~\ref{alg:common}, lns.~5--8), and apply a mixture model \citep{dempster1977maximum} (Alg.~\ref{alg:common}, ln.~12) to separate them into three distinct categories.

\begin{enumerate}
    \item low redundancy \textit{i.e.} cosine similarity near zero
    \item medium redundancy \textit{i.e.} cosine similarity $\approx0.5-0.8$
    \item high redundancy \textit{i.e.} cosine similarity $>0.9$
\end{enumerate}

\noindent Very few layers (empirically, $<15\%$) fall into the the low and medium redundancy categories, so we select all of these layers and for simplicity, and retain all of their attention heads. The remaining layers will be used for personalized attention head selection via Algo.~\ref{alg:individual}. These two disjoint subsets are the end goal of Algo.~\ref{alg:common} where they are defined as $\mathcal{L}_{\text{keep}}$ and $\mathcal{L}_{\text{drop}}$, respectively. Common layers serve two purposes: preserve the functionality of the most important layers, and provide a shared structural backbone between all clients to defend against client drift.
\vspace{-10pt}

\subsubsection{Individual Head Selection}
\label{sec:individual}
\begin{algorithm}
\caption{SelectIndividualHeads}
\label{alg:individual}
\begin{algorithmic}[1]
\Require $\mathcal{C}$, $\{s_{c,\ell,h}\}$, $\mathcal{L}_{\text{drop}}$ (from Alg.~\ref{alg:common}), $\lambda$, $\{N_\ell\}$, $|H_\ell|$
\State Initialize $\mathbf{Q} \gets \{ c \mapsto \emptyset \ \forall c \in \mathcal{C} \}$ \label{line:init-output}
\For{$\ell \in \mathcal{L}_{\text{drop}}$} \label{line:layer-loop}
    \State $\mu \gets |\mathcal{C}| N_\ell / |H_\ell|$ \label{line:mu}
    \State Solve for $x_{c,h} \in \{0,1\}$ and $u_h, v_h \ge 0$: \label{line:opt}
    \State \quad $\max \sum_{c, h} s_{c,\ell,h} x_{c,h} - \lambda \sum_{h} (u_h + v_h)$
    \State \quad s.t. $\sum_{c} x_{c,h} - \mu \le u_h, \quad \forall h$
    \State \quad \phantom{s.t.} $\mu - \sum_{c} x_{c,h} \le v_h, \quad \forall h$
    \State \quad \phantom{s.t.} $\sum_{h} x_{c,h} = N_\ell, \quad \forall c$
    \ForAll{$c \in \mathcal{C}$} \label{line:client-loop}
        \State $\mathbf{Q}[c][\ell] \gets \{ h \mid x_{c,h} = 1 \}$ \label{line:assign}
    \EndFor
\EndFor
\Return $\mathbf{Q}$
\end{algorithmic}
\end{algorithm}
\vspace{-4pt}

Following a consensus on common layers, we must determine how to personalize the selection of attention heads within the remaining layers $\mathcal{L_{drop}}$, with respect to each client's local data. Common methods for determining parameter significance in neural networks include magnitude-based scoring of individual weights \citep{franklelottery}, gradient-based importance measures using gradient magnitude as an indicator of a parameter's influence on model output \citep{wang2023label, panos2025efficient}, and higher-precision methods that use second-order derivatives to examine the curvature of the loss landscape \citep{lecun1989brain}. We select gradient-based scoring because it provides the most information about the importance of the attention head for a specific client's final representations, without the computational overhead of second-order methods \citep{agarwal2017second}. Specifically, we adopt the scoring metric of \citet{panos2025efficient} who fine-tune Vision-Language Models (VLM)  on domain specific datasets by targeting the most informative attention heads for localized parameter updates. To identify these attention heads, they extract the gradients of the loss function with respect to parameters $\mathbf{W}_{qh}$, $\mathbf{W}_{kh}$ and $\mathbf{W}_{vh}$, we denote these as $\mathbf{G}_{qh}$, $\mathbf{G}_{kh}$ and $\mathbf{G}_{vh}$, respectively. They then define an importance score ($s_h$) for each head, as shown in Eq.~\ref{eq:panos_score}.

\begin{equation} \label{eq:panos_score}
s_h = \sum_{i, j} (\mathbf{G}_{qh})_{i, j}^2 + (\mathbf{G}_{kh})_{i,j}^2 + (\mathbf{G}_{vh})_{i,j}^2
\end{equation}

\noindent During the course of our experiments, we discover that the $(\mathbf{G}_{vh})_{i,j}^2$ term dominates (see Appendix) so we simplify Eq.~\ref{eq:panos_score}, as shown in Eq.~\ref{eq:score} and adopt this simplified score as our head scoring metric where we denote the score for head $h$ of layer $\ell$ for client $c$ as $s_{c, \ell, h}$

\begin{equation} \label{eq:score}
s_h = \sum_{i, j} (\mathbf{G}_{vh})_{i,j}^2
\end{equation}

\noindent Previous work has demonstrated that model performance suffers when parameters are not trained evenly \citep{alam2022fedrolex, tianchi2025fedbrb, babakniya2023slora}. This constraint rules out the naive solution in which each client simply retains their top-k scoring heads without considering the choices of the others. In what follows, we formalize the constraints for attention head selection in one layer, $\ell$, across all clients, $c\in C$, for the selection of $N_\ell$ attention heads out of the total set of attention heads, $H_\ell$ in layer $\ell$.
\vspace{-10pt}

\begin{align}
    &\text{Let } x_{c, \ell, h}=
    \begin{cases}
        1 & \text{head }h \text{ is assigned to client } c \\
        0 & \text{otherwise} \\
    \end{cases} \nonumber \\
    &\text{Let } \sum_{h=1}^{H_\ell} x_{c,\ell, h}=N_\ell\;\forall c \nonumber \\
     &\text{Let } \mu = \frac{|C|N_\ell}{|H_\ell|} \nonumber \\
     &\lfloor\mu\rfloor\leq\sum_{c=1}^C x_{c,\ell, h}\leq \lceil\mu\rceil\;\forall h \label{eq:hard_bounds}  
\end{align}

\noindent Eq.~\ref{eq:hard_bounds} creates hard bounds, enforcing a strictly even distribution of local parameters between clients, a constraint widely accepted as necessary  based on previous work \citep{alam2022fedrolex, tianchi2025fedbrb}; however, these methods were developed for pre-training and federated fine-tuning behaviour can differ considerably \citep{nguyenbegin, legate2023guiding}. We hypothesize that pre-trained model weights, particularly those of large foundation models with strong initial representations, may show more tolerance when it comes to the necessity of even parameter training between clients, so we relax these hard bounds with slack variables $u_h$ and $v_h$ where $\sum_c x_{c, h}-\mu\leq u_h$ and $\mu-\sum_c x_{c, h}\leq v_h$ and use them to define a more flexible objective function in Eq.~\ref{eq:head_assign_objective}.

\begin{equation} \label{eq:head_assign_objective}
    \max\sum_{c=1}^C\sum_{h=1}^{H_\ell} s_{c,\ell, h}x_{c, \ell, h}-\lambda\sum_{h=1}^{H_\ell}(u_h+v_h) 
\end{equation}

\noindent Eq.~\ref{eq:head_assign_objective} optimizes head assignments by taking into account per-client head scores defined in Eq.~\ref{eq:score} and enforces balanced assignments according to hyperparameter $\lambda$ which governs how strictly balanced parameter training should be enforced (lns. 4-8 of Algo.~\ref{alg:individual}). The result of Algo.~\ref{alg:individual} is $\mathbf{Q}$, a data structure that contains attention head assignments for each client, $c\in C$, for each layer $\ell\in L$. These assignments provide the recipe to create each client's unique, data-informed Hydra-Net.

\section{Experiments} \label{sec:exp}
\begin{figure*}[!t]
\centering
\setlength{\tabcolsep}{1.5pt}
\renewcommand{\arraystretch}{0.85}

\begin{tabular}{ccccc}

    \includegraphics[width=0.19\textwidth]{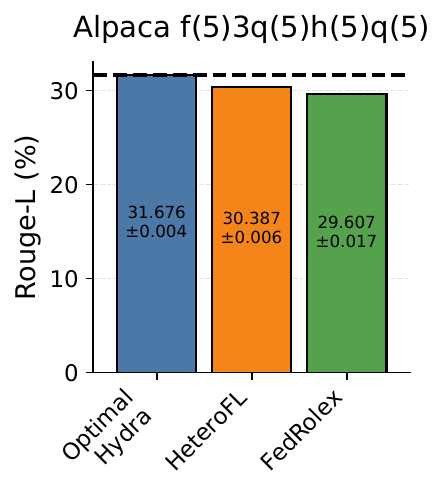} &
    \includegraphics[width=0.19\textwidth]{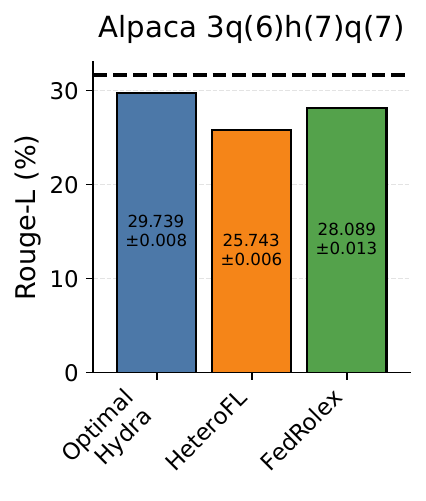} &
    \includegraphics[width=0.19\textwidth]{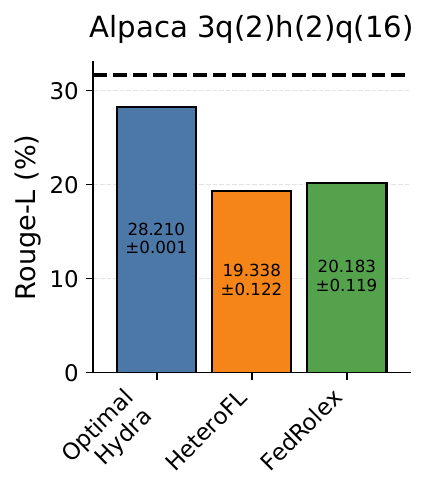} &
    \includegraphics[width=0.19\textwidth]{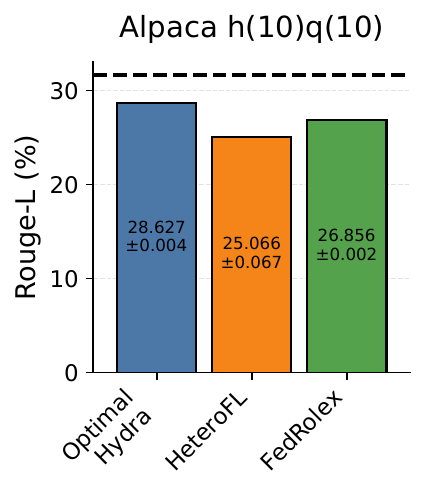} &
    \includegraphics[width=0.19\textwidth]{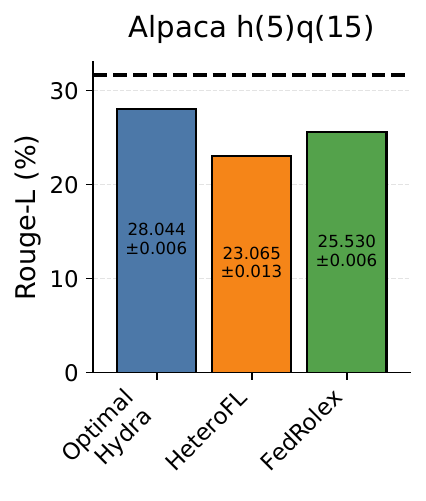}
    \\[-2pt]

    \includegraphics[width=0.19\textwidth]{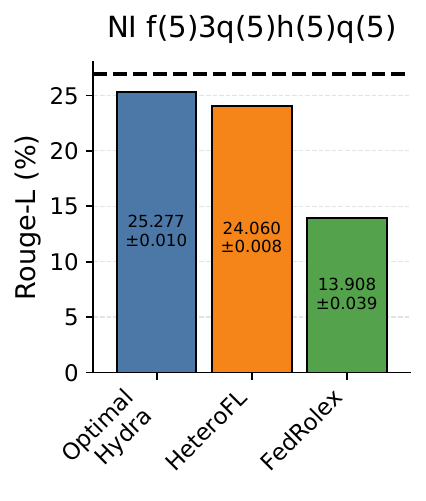} &
    \includegraphics[width=0.19\textwidth]{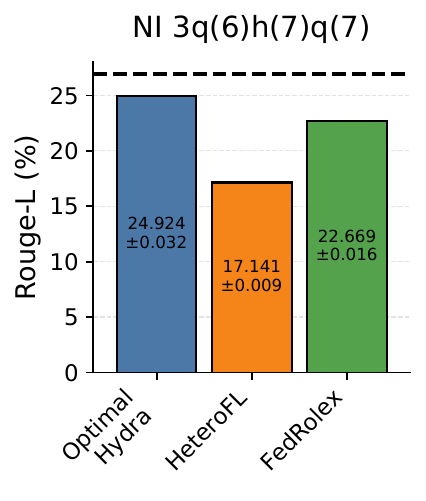} &
    \includegraphics[width=0.19\textwidth]{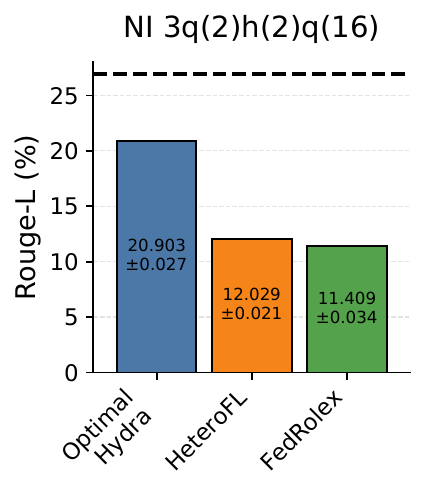} &
    \includegraphics[width=0.19\textwidth]{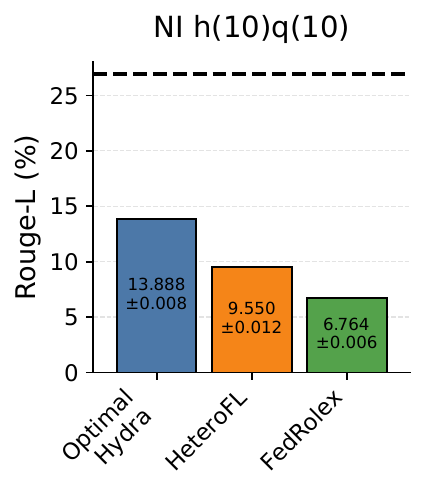} &
    \includegraphics[width=0.19\textwidth]{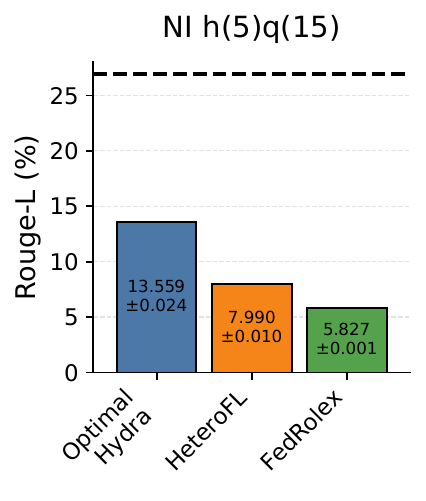}
    \\[-2pt]

    \includegraphics[width=0.19\textwidth]{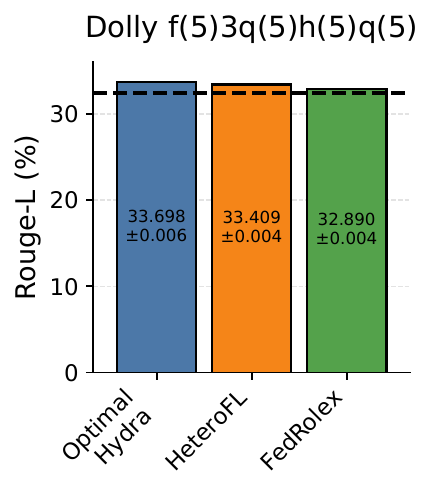} &
    \includegraphics[width=0.19\textwidth]{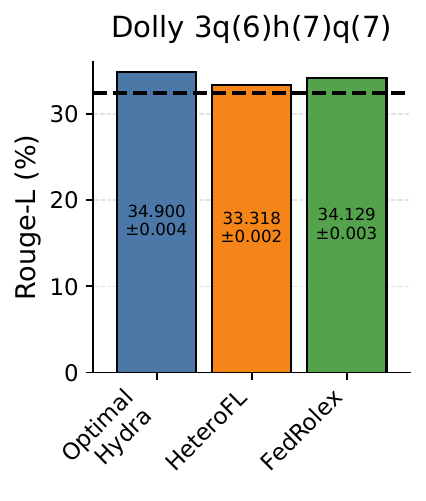} &
    \includegraphics[width=0.19\textwidth]{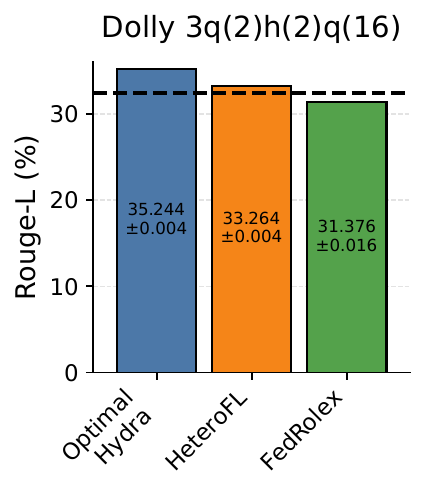} &
    \includegraphics[width=0.19\textwidth]{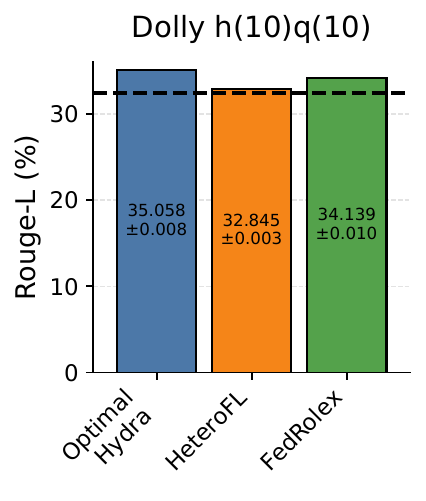} &
    \includegraphics[width=0.19\textwidth]{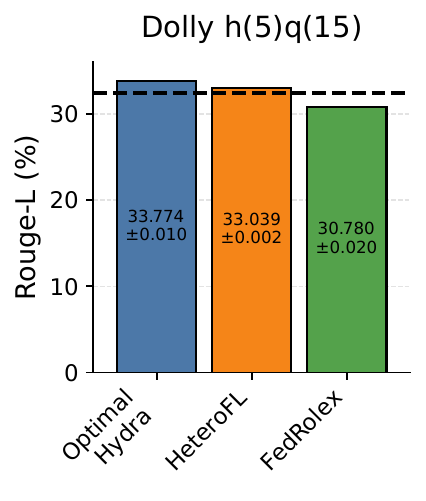}

\end{tabular}
\vspace{-10pt}
\caption{Each row consists of results for one dataset (1. Alpaca 2. Natural Instructions 3. Dolly 15K) and each column corresponds to one model heterogeneous FL configuration (1. \textsc{F(5)3Q(5)H(5)Q(5)} 2. \textsc{3Q(6)H(7)Q(7)} 3. \textsc{3Q(2)H(2)Q(16)} 4. \textsc{H(10)Q(10)} 5. \textsc{H(5)Q(15)}). Each sub-figure contains the mean Rouge-L $\pm$ std. dev. of four seeds for Optimal Hydra (blue), and our transformer adapted HeteroFL (orange) and FedRolex (green) baselines. The black dotted line indicates the FedAvg baseline whose values are reported in Tab.\ref{tab:homo-main}. Optimal Hydra performs better than the baselines in all cases, it does particularly well with the very data heterogeneous NI dataset and the very resource constrained \textsc{3Q(2)H(2)Q(16)} model heterogeneous configuration.\vspace{-12pt}}
\label{fig:hetero-main}
\end{figure*} 

While Optimal Hydra is feasible for any number of attention heads dropped, we study model heterogeneity by focusing on four discreet sizes; all attention heads kept, or full (\textsc{f}), three quarters of attention heads kept (\textsc{3q}), half of attention heads kept (\textsc{h}), one quarter of attention heads kept (\textsc{q}). Heterogeneous configurations are indicated by \textsc{f}, \textsc{3q}, \textsc{h}, \textsc{q}, followed by the number of clients assigned that model size in brackets.

\paragraph{Datasets} We consider three datasets, Natural Instructions (NI) \citep{natural_instructions}, Alpaca \citep{alpaca} and Dolly-15K \citep{DatabricksBlog2023DollyV2}. Dolly-15K samples are distributed between clients using a Dirichlet distribution with $\alpha=0.5$. The creative writing task of Dolly-15K is not included in any of the local client data distributions and is instead used for global evaluation. The Alpaca dataset does not include pre-determined task categories, so we define six categories, summarization, reasoning, classification, mathematics, QA and other, and assign each sample a classification by searching for keywords in the input. As with Dolly-15K, samples are distributed between clients using a Dirichlet distribution with $\alpha=0.5$. The``other" task is selected for global evaluation. For NI, which has 857 unique tasks, 119 test tasks are held-out and used for global evaluation, while each of the remaining 738 tasks is uniquely assigned to one client. This isolated task distribution makes the NI dataset particularly challenging.

\paragraph{Training Details}
We use the DataJuicer-1.3 model, a Llama-1.3B model pre-trained on an open source corpus of data  \citep{chen2024data} and the Rouge-L \citep{lin-2004-rouge} metric to evaluate fine-tuned model performance, reported scores are the mean of four seeds. Each experiment has 20 clients and all of them participate fully in each round of training.  Since and some distributions may be more amenable than others to optimal sub-net selection, client-side data distributions are important variables to control.  To remove any ambiguity that may result from local data variation, we manually set a seed for each experiment to ensure that each client's local data remains the same for each set of experiments done on each model heterogeneity configuration. Additional training details and hyper-parameter settings are provided in the Appendix.


\paragraph{Baselines}
Since there are no pre-existing model heterogeneous FL transformer architectures in the literature, we make appropriate transformer based adaptations of \textit{HeteroFL} and \textit{FedRolex} to compare against Optimal Hydra. We additionally evaluate Optimal Hydra against the zero-shot performance of the pre-trained model and full federated training using FedAvg \citep{mcmahan2017communication}.


\noindent \textit{Adapted HeteroFL and FedRolex:}
HeteroFL \citep{diaoheterofl} trains static sub-networks with parameters distributed un-equally between clients. Since Optimal Hydra is focused on attention head selection, the adapted HeteroFL baseline retains the first $N_\ell$ heads in each layer. FedRolex \citep{alam2022fedrolex} rotates consecutive sub-network parameters for clients after each round of training. To adapt FedRolex we select $N_\ell$ consecutive attention heads starting at a random head index in each layer. Continuity is achieved by rolling indices so that, following head index $N$, the next consecutive index is $0$. The initial head index for each layer is incremented by 1 after each round, and client sub-networks are updated accordingly.

\paragraph{Improved Resource Utilization:}
\begin{table}[!t]
\centering
\small 
\setlength{\tabcolsep}{2.5pt} 

\caption{Memory and compute requirements for the different resource allocation configurations we study. We provide both the absolute value and a percentage improvement over full FedAvg fine-tuning. Configurations are ordered from most to least resource intensive.}
\label{tab:resources}
\vspace{5pt}
\begin{tabular}{@{} l c c c c @{}}
    \toprule
    \multirow{2}{*}{\textsc{Config}} 
    & \textsc{Comp.} 
    & \textsc{\% Impr.} 
    & \textsc{Mem.} 
    & \textsc{\% Impr.} \\ 
    & \textsc{(GFLOPS)} 
    & \textsc{w.r.t. full} 
    & \textsc{(GB)} 
    & \textsc{w.r.t. full} \\
    \midrule
    \textsc{f(20)}             & 56.6 & --   & 61.9 & --   \\
    \textsc{3q(20)}            & 50.6 & 10.7 & 55.8 & 9.7  \\
    \textsc{f(5)3q(5)h(5)q(5)} & 47.6 & 16.0 & 52.8 & 14.6 \\
    \textsc{h(20)}             & 44.5 & 21.3 & 49.8 & 19.5 \\
    \textsc{3q(6)h(7)q(7)}     & 44.2 & 21.9 & 49.5 & 20.0 \\
    \textsc{h(10)q(10)}        & 41.5 & 26.7 & 46.8 & 24.4 \\
    \textsc{3q(2)h(2)q(16)}    & 40.3 & 28.8 & 45.6 & 26.4 \\
    \textsc{h(5)q(15)}         & 40.0 & 29.3 & 45.3 & 26.8 \\
    \textsc{q(20)}             & 38.5 & 32.0 & 43.7 & 29.3 \\
    \bottomrule
\end{tabular}
\vspace{-10pt}
\end{table}

\begin{table*}[t]
\centering
\footnotesize
\setlength{\tabcolsep}{4pt}

\caption{Rouge-L (\%) $\pm$ the std. dev. for four seeds, is reported for The full model zero-shot and FedAvg baselines as well as the setting in which all clients have homogeneous, but limited resource capabilities.}
\label{tab:homo-main}
\vspace{5pt}
\begin{tabular}{@{} l ccc ccc ccc @{}}
    \toprule
    & \multicolumn{3}{c}{\textsc{Alpaca}}
    & \multicolumn{3}{c}{\textsc{Natural Instructions}}
    & \multicolumn{3}{c}{\textsc{Dolly}} \\
    \cmidrule(lr){2-4}
    \cmidrule(lr){5-7}
    \cmidrule(lr){8-10}
    & \textsc{Hydra} & \textsc{HeteroFL} & \textsc{FedRolex}
    & \textsc{Hydra} & \textsc{HeteroFL} & \textsc{FedRolex}
    & \textsc{Hydra} & \textsc{HeteroFL} & \textsc{FedRolex} \\
    \midrule
    \textsc{F(20) ZS}
        & \multicolumn{3}{c}{$17.157$}
        & \multicolumn{3}{c}{$5.698$}
        & \multicolumn{3}{c}{$26.694$} \\
    \addlinespace
    
    \textsc{F(20) FedAvg}
        & \multicolumn{3}{c}{$31.668 \ (\pm.002)$}
        & \multicolumn{3}{c}{$26.915 \ (\pm.002)$}
        & \multicolumn{3}{c}{$32.465 \ (\pm.001)$} \\
    \midrule
    
    \textsc{3Q(20)}
        & $\mathbf{31.267}$ & $26.183$ & $30.186$
        & $25.593$ &  $22.991$ &  $\mathbf{25.630}$
        & $34.295$ & $29.535$ & $\mathbf{34.714}$ \\
        & $\boldsymbol{(\pm.001)}$ & $(\pm.002)$ & $(\pm.001)$
        & $(\pm.005)$ & $(\pm.005)$ & $\boldsymbol{(\pm.004)}$
        & $(\pm.005)$ & $(\pm.002)$ & $\boldsymbol{(\pm.003)}$ \\
    \addlinespace
    
    \textsc{H(20)}
        & $\mathbf{30.621}$ & $24.077$ & $29.258$
        & $\mathbf{22.738}$ &  $17.821$ & $22.157$
        & $\mathbf{35.489}$ & $31.516$ & $35.431$ \\
        & $\boldsymbol{(\pm.005)}$ & $(\pm.003)$ & $(\pm.005)$
        & $\boldsymbol{(\pm.010)}$ & $(\pm.012)$ & $(\pm.004)$
        & $\boldsymbol{(\pm.002)}$ & $(\pm.001)$ & $(\pm.002)$ \\
    \addlinespace
    
    \textsc{Q(20)}
        & $\mathbf{25.232}$ & $8.264$  & $15.224$
        & $7.990$ &  $\mathbf{8.077}$ & $7.609$
        & $\mathbf{29.104}$ & $28.881$ & $20.552$ \\
        & $\boldsymbol{(\pm.008)}$ & $(\pm.080)$ & $(\pm.129)$
        & $(\pm.006)$ & $\boldsymbol{(\pm.002)}$ & $(\pm.004)$
        & $\boldsymbol{(\pm.007)}$ & $(\pm.002)$ & $(\pm.027)$ \\
    \bottomrule
\end{tabular}
\vspace{-10pt}
\end{table*}

\noindent Tab.~\ref{tab:resources} provides memory and compute comparisons for the resource allocation configurations in each of our experiments. These values are also expressed as the percentage improvement each configuration provides over the full FedAvg baseline. In the most extreme case where each client selects only $25\%$ of the attention heads, we achieve savings of $32.0\%$ for computation and $29.3\%$ for memory. Values provided in Tab.~\ref{tab:resources} are calculated using a sequence length of 2048; however, since computational requirements depend heavily on sequence length, we note that an increase in sequence length to $\approx 12 000$ achieves $50\%$ savings in computational resources. This computational sequence length dependence theoretically makes Optimal Hydra an attractive prospect for FL with long context LLMs.

\paragraph{Results:}
The results in Tab.~\ref{tab:homo-main} consider a homogeneous resource setting in which all clients retain the same number of attention heads, but may retain different subsets of heads. In this scenario, clients have equal resource capabilities, although these budgets may still be substantially smaller than those required for full-model fine-tuning.  We also report the zero-shot performance of the pre-trained model and the performance of full-model FedAvg.

\noindent In this setting, the relative performance of the methods is mixed. Optimal Hydra achieves the best score in the majority of configurations and is consistently the best-performing method or competitive with the strongest baseline. The adapted HeteroFL and FedRolex baselines exhibit less consistency; HeteroFL performs substantially worse than the other methods for \textsc{H(20)} on Alpaca, and FedRolex exhibits a similar sub-standard performance for \textsc{Q(20)} on Dolly-15K. These results suggest that, when all clients have the same resource budget, the benefit of data-informed head selection is present but relatively modest, and the choice of which heads are retained becomes increasingly important as the available model capacity decreases.

\noindent Fig.~\ref{fig:hetero-main} considers a more challenging setting in which clients have substantially different resource constraints. We assign different fractions of clients different model sizes, with particular emphasis on configurations where most clients are under-resourced. Unlike the homogeneous-resource setting, Optimal Hydra consistently outperforms both baselines across all heterogeneous resource configurations. The advantage is most pronounced for the NI dataset, which has the most heterogeneous task distribution and supports our hypothesis that sub-network selection should account for both the resource constraints and client data distributions. Optimal Hydra also performs particularly well in the aforementioned low-resource setting where it achieves a $42.5\%$ improvement over FedRolex, the next highest scoring baseline, on the \textsc{3Q(2)H(2)Q(16)} configuration for the NI dataset.

 \noindent Finally, model heterogeneity does not necessarily result in a performance degradation relative to full-model FedAvg. We observe that Optimal Hydra and the adapted baselines sometimes outperform the full-model FedAvg baseline, one possible explanation is that removing attention heads acts as a form of regularization during fine-tuning. This effect is most pronounced on Dolly-15K where the pre-trained model already provides strong initial representations. 

 \paragraph{Equal Parameter Training in the Fine-Tuning Setting:}
 Previous work in the pre-training setting has motivated the importance of evenly training model parameters \citep{alam2022fedrolex, tianchi2025fedbrb}; however, application of this principle to fine-tuning has not been closely examined, despite indications of behavioural differences between these two regimes \citep{legate2023guiding, nguyenbegin}. The objective in Eq.~\ref{eq:head_assign_objective} introduces an additional hyperparameter, $\lambda$, which controls how strictly equal parameter training is enforced. When $\lambda=0$, the balance penalty vanishes and head assignments are determined solely by client-specific importance scores. As $\lambda$ increases, the objective increasingly favors equal head distribution across clients, $\lambda\approx 1\times10^6$ effectively recovers the hard-bounded objective in Eq.~\ref{eq:hard_bounds}, enforcing an even distribution of heads between clients for each layer.

\noindent Dolly-15K has the lowest attention head score diversity between clients, with clients selecting similar subsets of heads in each layer. The best Rouge-L scores were often achieved when $\lambda=0$ or $\lambda=1$, indicating that the strict requirement for equal parameter training established in federated pre-training may be greatly relaxed during fine-tuning. NI displayed the greatest attention head score diversity, with clients naturally gravitating towards different attention heads. NI generally achieved better Rouge-L scores with larger values of $\lambda$, sometimes as high as 500 or 1000, indicating a greater need for enforced attention head diversity.

\noindent To account for the difference in optimal $\lambda$ values, we turn to the zero-shot performance strength of the pretrained representations. Dolly-15K had zero-shot performance of $26.694$, significantly higher than the $5.698$ achieved on NI. The strong initial representations available to Dolly-15K make strictly even parameter fine-tuning unnecessary. The weaker initial representations available for NI may make greater participation of all parameters more important, resulting in behavior more similar to the equal parameter training observed in the pre-training regime.

\section{Conclusion}
We introduced Optimal Hydra, a model-heterogeneous federated learning framework for fine-tuning large language models under heterogeneous client resource constraints. Optimal Hydra creates data-informed Hydra-Nets that permit resource limited clients to adapt to their available resources, by retaining model parameters best suited to their own local data. Across three datasets and multiple resource configurations, Optimal Hydra consistently outperformed adapted model-heterogeneous FL baselines in the most resource-heterogeneous settings while reducing local computational and memory requirements by up to $32.0\%$ and $29.3\%$, respectively. Our results also challenge the assumption that model parameters must be trained equally across clients. In the fine-tuning setting, the necessary degree of parameter balancing depends on the strength of the pretrained representations and the heterogeneity of the local data distributions.


\bibliography{main}
\bibliographystyle{tmlr}

\appendix
\section{Memory and Compute Calculations} \label{apx: comp_mem}
This section develops the rationale behind the values presented in the Improved Resource Utilization part of the Experiments section in the main body of the paper. Table~\ref{tab:var_defs} summarizes the variables and values used in the calculations in sections \ref{apx:comp} and \ref{apx:mem}:

\begin{table}[h!]
    \centering
    \begin{tabular}{|c|c|c|}
        &&\\
        \toprule
        \textbf{Name} & \textbf{Var.} & \textbf{DataJuicer-1.3B} \\
        \midrule
        No. trf blocks 1 & $L$ &  24 \\
        Hidden dim. &$d$& 2048 \\
        Tot. No. heads &$h$& 16 \\
        Reduced No. heads &$h'$& variable \\
        Head dim. & $d_h$ & 128 \\
        FF dim. & $i$ &  5504 \\
        Vocab. size & $v$  & 32000 \\
        Seq. Len. & $s$ & 2048\\
        \bottomrule
    \end{tabular}
    \caption{Provides a summary of variables and their values for the DataJuicer-1.3B model architectures}
     \label{tab:var_defs}
\end{table}
\subsection{Compute}\label{apx:comp}
Here we adopt the convention that matrix multiplication operations of the form $(M\times K)\cdot (K\times N)$ require 2MKN flops. We define a scaling factor $\alpha$ to account for the reduction in attention heads $ s.t.\; \alpha=\frac{h'}{h}$. 

\paragraph{Attention Block}
Attention in each transformer block consists of weights $\mathbf{W}_q$, $\mathbf{W}_k$, $\mathbf{W}_v$ and $\mathbf{W}_o$, each with dimension $(d\times d)$. 
\begin{itemize}
    \item \textbf{Linear Proj:} Accounting for $\mathbf{W}_q$, $\mathbf{W}_k$, $\mathbf{W}_v$ and $\mathbf{W}_o$ with scaling for head reduction, we get:  \textcolor{blue}{$8d^2\alpha$}
    \item \textbf{Score:} $QK^T$ multiplies with the KV cache for each head: \textcolor{blue}{$2sd \alpha$}
     \item \textbf{Value Aggregation:} Similar ops to score: \textcolor{blue}{$2sd \alpha$}
\end{itemize}
\textcolor{red}{\textbf{Attention Total (per layer):} $8d^2\alpha + 4sd\alpha$}

\paragraph{Feed Forward Block} Llama used SwiGLU \cite{shazeer2020glu} which has weights $\mathbf{W}_{gate}$, $\mathbf{W}_{up}$, both with dimension $(d\times i)$ and $\mathbf{W}_{down}$ with dimension $(i\times h)$. Operations for each set of weights cost \textcolor{blue}{$2di$} flops.\\

\textcolor{red}{\textbf{FF Total (per layer):} $6di$}

\paragraph{Output} map final hidden state back to the vocabulary. This is a matrix multiplication $(1\times d)\cdot (d\times v)$.\\

\textcolor{red}{\textbf{Output Total (one time contribution):} $2dv$}\\

Combining the contributions of these operations over $L$ layers, we get the following formula for the number of flops required for one forward pass.
\begin{equation}
    comp = L(8d^2\alpha + 4sd\alpha + 6di) + 2dv
\end{equation}

Since many of these values are constant for a given model architecture, we get the following expressions as a function of $\alpha$ and sequence length, $s$ for DataJuicer-1.3B.

\begin{align}
\textbf{DataJuicer-1.3B}&\nonumber\\
comp&=L(8d^2\alpha + 4sd\alpha + 6di) + 2dv \nonumber\\
&\approx24(8\times2048^2\alpha + 4\times2048 s\alpha + 6\times 2048 \times 5504) + 2\times 2048 \times 32000 \nonumber\\
&\approx\underbrace{8.05\times10^8\alpha + 1.97\times10^5s\alpha}_\text{Attention} + \underbrace{1.62\times10^9}_\text{FF (constant)} + \underbrace{1.31\times 10^8}_\text{output (constant)}\nonumber\\
&\approx4.295\times10^9\alpha + 5.24\times10^5s\alpha + 1.65\times10^9
\end{align}
 
\subsection{Memory} \label{apx:mem}
For memory calculations we assume 16-bit precision giving us 2 bytes per parameter. Memory requirements can be divided into two categories: static memory resulting from model weights and dynamic memory resulting from the KV-cache.

\paragraph{Static Memory}
Using the previously provided weight dimensions and $\alpha$ scaling factor, we can compute static memory requirements as follows
\begin{itemize}
    \item \textbf{Attention:} Weights $\mathbf{W}_q$, $\mathbf{W}_k$, $\mathbf{W}_v$ and $\mathbf{W}_o$ account for:  \textcolor{blue}{$4d^2\alpha\times 2\;\text{bytes}$} per layer
    \item \textbf{FF:} Weights $\mathbf{W}_{gate}$, $\mathbf{W}_{up}$, and $\mathbf{W}_{down}$ account for:  \textcolor{blue}{$3di\times 2\;\text{bytes}$} per layer
     \item \textbf{Output/Embedding:} accounts for:  \textcolor{blue}{$2vd\times 2\;\text{bytes}$}
\end{itemize}
\textcolor{red}{\textbf{Total Static Memory:} $L(8d^2\alpha + 6di) + 4vd$}

\paragraph{Dynamic Memory} The KV-cache contribution is \textcolor{blue}{$\underbrace{2}_\text{1 key, 1 val.} \times s(h'\times d_h)\times 2\;\text{bytes}$} per layer.

\textcolor{red}{\textbf{Total Dynamic Memory:} $4Ls (h'd_h)$}

Combining the contributions of static and dynamic memory, we get the following formula for the number of bytes required for one forward pass.
\begin{equation}
    mem = L(8d^2\alpha + 6di) + 4vd + 4Lsh\alpha
\end{equation}

Again, many of these values are constant for a given model architecture, so we get the following expressions as a function of $\alpha$ and sequence length, $s$ for DataJuicer-1.3B.

\begin{align}
\textbf{DataJuicer-1.3B}&\nonumber\\
mem&= L(8d^2\alpha + 6di) + 4vd + 4Lsh\alpha \nonumber\\
&\approx24(8\times2048^2\times\alpha + 6\times2048\times5504) + 4\times 32000\times2048 + 4\times24\times s\times 2048\times\alpha \nonumber\\
&\approx\underbrace{8.05\times10^8\alpha}_\text{static memory (attention)} + \underbrace{1.62\times10^9}_\text{static memory (FF)} + \underbrace{2.62\times10^8}_\text{static memory (optput)} + \underbrace{1.97\times10^5s\alpha}_\text{dynamic memory}\nonumber\\
&\approx8.05\times10^8\alpha + 1.89\times10^9 + 1.97\times10^5s\alpha
\end{align}
\vspace{15pt}

\section{Experimental Settings}
Experiments were run across multiple compute clusters using a variety of GPUs including L4OS (48GB), RTX8000 (48GB), A100L (80GB), A100 (40GB), V100 (32GB), A6000 (48GB), H100 (80GB). Each experiment was executed using 2 or 4 GPUs, depending on the compute availability.
\vspace{15pt}

\section{Hyper-parameter Searches}
Hyperparameter tuning was conducted using a grid search over the relevant hyper-parameters. Federated learning has numerous tunable parameters including server and client learning rates, $\eta_s$ and $\eta_c$, respectively, the number of gradient steps (\textit{i.e.} iterations) taken prior to client aggregation, $I$, and the number of federated rounds, $R$. Optimal Hydra additionally requires tuning of hyper-parameter $\lambda$ that governs the degree to which equal parameter training is enforced.

\noindent We first specify hyper-parameters that are common for all datasets and heterogeneity configurations:
\begin{itemize}
    \item $R=40$
    \item $I=200$
    \item $\eta_s =1$
\end{itemize}

$\eta_c$ and $\lambda$ are tuned individually per method, dataset and heterogeneity combination, grid search ranges were specified by dataset as follows:
\begin{itemize}
    \item Alpaca: $\eta_c = \{5\times 10^3, 1\times 10^3, 5\times 10^4, 1\times 10^4\}$,  $\lambda = \{5, 10, 25, 50, 100, 500, 1000\}$
    \item Natural Instructions: $\eta_c = \{1\times 10^3, 7\times 10^4, 5\times 10^4, 1\times 10^4, 5\times 10^5, 1\times 10^5\}$,  $\lambda = \{50, 100, 500, 1000, 5000, 10000\}$
    \item Dolly 15K: $\eta_c = \{5\times 10^4, 1\times 10^4, 5\times 10^5, 1\times 10^5\}$,  $\lambda = \{0, 1, 5, 10, 25, 100, 500, 1000\}$
\end{itemize}

\noindent Tab.~\ref{tab:hps} specifies the final hyper-parameter combinations used for each result, 

\begin{table}[h!]
    \centering
    \begin{tabular}{|l|c|c|c|c|c|}
    &&&&&\\
    &&&&&\\
        \toprule
        \textbf{Dataset} & \textbf{Config} & $\eta_c$ \textsc{(OH)} & $\lambda$ \textsc{(OH)} &$\eta_c $ \textsc{(HFL)}&$\eta_c$ \textsc{(Rolex)}\\
        \midrule
        \multirow{8}{*}{Alpaca} 
        & \textsc{3Q(20)} &$5\times 10^3$  &$100$ & $5\times 10^4$& $1\times 10^3$ \\
        & \textsc{H(20)} &$5\times 10^3$  & $10$ & $5\times 10^4$ &$5\times 10^4$\\
        & \textsc{Q(20)} & $1\times 10^3$ & $100$ & $5\times 10^4$ &$5\times 10^4$\\
        & \textsc{F(5)3Q(5)H(5)Q(5)} & $5\times 10^3$ & 25 & $1\times 10^3$ & $1\times 10^3$\\
        & \textsc{H(10)Q(10)} & $1\times 10^3$ & $25$ & $1\times 10^3$ & $1\times 10^4$\\
        & \textsc{3Q(6)H(7)Q(7)} & $5\times 10^3$ & $50$ & $5\times 10^4$& $5\times 10^3$\\
        & \textsc{H(5)Q(15)} & $1\times 10^3$ & $25$ & $5\times 10^4$& $5\times 10^4$  \\
        & \textsc{3Q(2)H(2)Q(16)} & $5\times 10^3$ & $100$& $5\times 10^4$& $5\times 10^4$ \\
        \cmidrule{1-6}
        \multirow{8}{*}{Dolly 15K} 
        & \textsc{3Q(20)} & $5\times 10^5$ & $0$ & $5\times 10^5$& $1\times 10^5$\\
        & \textsc{H(20)} & $1\times 10^4$ & $100$ & $5\times 10^5$& $5\times 10^5$ \\
        & \textsc{Q(20)} & $1\times 10^5$ & $0$ & $1\times 10^5$& $5\times 10^6$\\
        & \textsc{F(5)3Q(5)H(5)Q(5)} & $1\times 10^5$ & $25$ &$5\times 10^5$ & $5\times 10^5$\\
        & \textsc{H(10)Q(10)} & $5\times 10^5$ & $50$ &$5\times 10^5$ & $5\times 10^5$\\
        & \textsc{3Q(6)H(7)Q(7)} & $5\times 10^5$ & $25$ &$5\times 10^5$& $5\times 10^5$\\
        & \textsc{H(5)Q(15)} & $5\times 10^5$ & $1$ &$5\times 10^5$&$5\times 10^5$\\
        & \textsc{3Q(2)H(2)Q(16)} & $5\times 10^5$ & $5$ & $5\times 10^5$& $5\times 10^5$\\
        \cmidrule{1-6}
        \multirow{8}{*}{NI} 
        & \textsc{3Q(20)} & $7\times 10^4$ & 500 & $5\times 10^4$ & $5\times 10^4$\\
        & \textsc{H(20)} & $5\times 10^4$ & 100 & $5\times 10^4$ & $1\times 10^4$ \\
        & \textsc{Q(20)} & $5\times 10^5$ & 5000 & $1\times 10^4$ & $5\times 10^5$ \\
        & \textsc{F(5)3Q(5)H(5)Q(5)} & $1\times 10^4$ & $100$ &$1\times 10^4$ & $1\times 10^5$\\
        & \textsc{H(10)Q(10)} &  $1\times 10^4$ & $100$ &$1\times 10^4$ & $5\times 10^5$  \\
        & \textsc{3Q(6)H(7)Q(7)} & $1\times 10^4$ & $100$ &$1\times 10^4$ & $1\times 10^4$  \\
        & \textsc{H(5)Q(15)} &  $1\times 10^4$ & $50$ &$1\times 10^4$ & $5\times 10^6$   \\
        & \textsc{3Q(2)H(2)Q(16)} & $1\times 10^4$ & $500$ &$1\times 10^4$ & $1\times 10^4$   \\
        \bottomrule
    \end{tabular}
    \caption{Provides selected hyper-parameters for Optimal Hydra each dataset, heterogeneity combination evaluated in the main body of the paper. OH stands for Optimal Hydra, HFL and Rolex for our modified HeteroFL and FedRolex baselines, respectively.}
    \label{tab:hps}
\end{table}

\vspace{15pt}

\section{$\nabla\mathbf{(W_v)^2}$ Dominates Head Scoring} \label{apx:qkv_score}
In the Individual Head Selection section of the main paper we reference an importance score ($s_h$) utilized in the determination of attention head importance by \citet{panos2025efficient}) (Eq.~\ref{eq:panos_score})

\begin{equation} \label{eq:panos_score}
s_h = \sum_{i, j} (\mathbf{G}_{qh})_{i, j}^2 + (\mathbf{G}_{kh})_{i,j}^2 + (\mathbf{G}_{vh})_{i,j}^2
\end{equation}

\noindent We justified our simplification of the scoring function to Eq.~\ref{eq:score} by claiming that in our experiments the $(\mathbf{G}_{vh})_{i,j}^2$ term dominates, the following plots have been selected to demonstrate this claim.

\begin{figure*}[h]
\hspace{160pt}
    \includegraphics[width=0.33\textwidth]{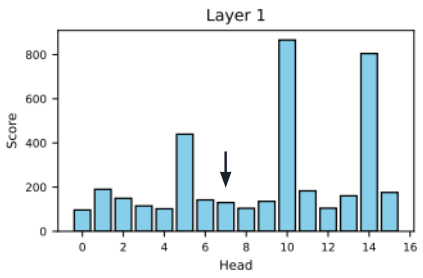}\\
    \includegraphics[width=0.33\textwidth]{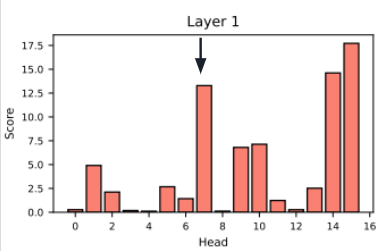}
     \includegraphics[width=0.33\textwidth]{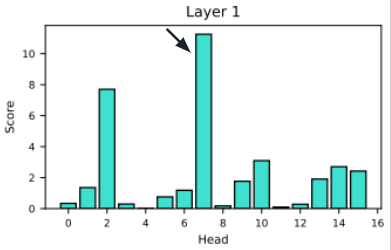}
      \includegraphics[width=0.33\textwidth]{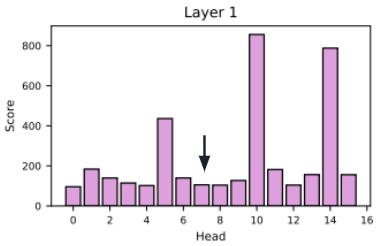}
    \caption{For Client 1 at round 20 and transformer layer 1, we see head 7 is unimportant to both $\nabla\mathbf{(W_v)^2}$ and $(\nabla \mathbf{W_q})^2 + (\nabla \mathbf{W_k})^2 + (\nabla \mathbf{W_v})^2$ and important to $(\nabla \mathbf{W_q})^2$ and $(\nabla \mathbf{W_k})^2$ showing the dominance of $\nabla\mathbf{(W_v)^2}$ in the scoring mechanism proposed in Eq.~\ref{eq:panos_score}}
     \label{fig:qkv1}
\end{figure*}

\begin{figure*}[h]
\hspace{160pt}
    \includegraphics[width=0.33\textwidth]{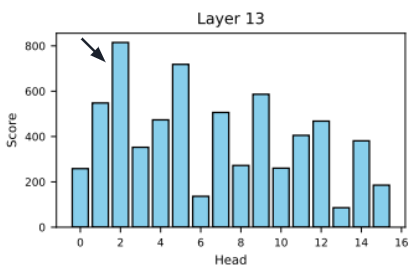}\\
    \includegraphics[width=0.33\textwidth]{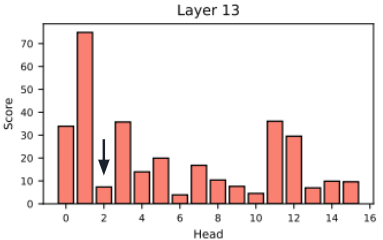}
     \includegraphics[width=0.33\textwidth]{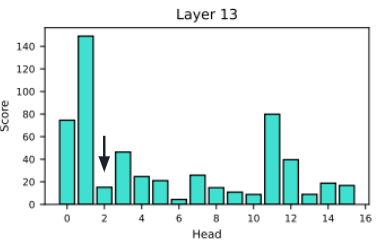}
      \includegraphics[width=0.33\textwidth]{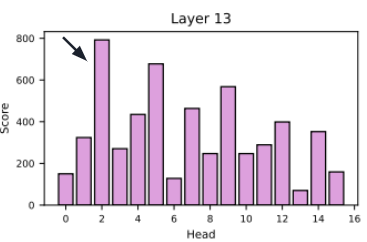}
    \caption{For Client 3 at round 5 and transformer layer 13, we see head 2 is important to both $\nabla\mathbf{(W_v)^2}$ and $(\nabla \mathbf{W_q})^2 + (\nabla \mathbf{W_k})^2 + (\nabla \mathbf{W_v})^2$ and unimportant to $(\nabla \mathbf{W_q})^2$ and $(\nabla \mathbf{W_k})^2$ again that $\mathbf{(W_v)^2}$ is dominant.}
    \label{fig:qkv13}
\end{figure*}

\noindent Figs.~\ref{fig:qkv1} and \ref{fig:qkv13} show selected instances of scores for a particular client, layer and training round to illustrate the dominance of $\nabla\mathbf{(W_v)^2}$. Plots show scores for $(\nabla \mathbf{W_q})^2$ (orange), $(\nabla \mathbf{W_k})^2$ (green), $(\nabla \mathbf{W_v})^2$ (purple) and $(\nabla \mathbf{W_q})^2 + (\nabla \mathbf{W_k})^2 + (\nabla \mathbf{W_v})^2$ (blue) were generated for federated rounds $\{1, 2, 3, 4, 5,15, 20, 30, 40\}$, for each transformer block and each client. Results indicate that the score from Eq.~\ref{eq:panos_score} is largely dominated by $(\nabla \mathbf{W_v})^2$. 

\end{document}